\documentclass[conference]{IEEEtran}
\IEEEoverridecommandlockouts
\usepackage{cite}
\usepackage{amsmath,amssymb,amsfonts}
\usepackage{algorithmic}
\usepackage{graphicx}
\usepackage{textcomp}
\usepackage{xcolor}
\usepackage[printonlyused,withpage,nolist,nohyperlinks]{acronym}
\usepackage[labelsep=quad,indention=10pt]{subfig}
\usepackage[export]{adjustbox}
\usepackage{bm}
\usepackage{comment}

\newcommand{\eg}{\textit{e}.\textit{g}., }

\begin{acronym} \renewcommand{\\}{}  
\acro{AdaNNMPC}{Adaptive Neural Network Model Predictive Control}
\acro{AUV}{Autonomous Underwater Vehicle}
\acro{COLA2}{Component Oriented Layer-based Architecture for Autonomy}
\acro{COTS}{commercial-off-the-shelf}
\acro{DoF}{Degree of Freedom}
\acro{DR}{Dead Reckoning}
\acro{DVL}{Doppler Velocity Log}
\acro{IMU}{Inertial Measurement Unit}
\acro{MPC}{Model Predictive Control}
\acro{OMPL}{Open Motion Planning Library}
\acro{OSL}{Ocean System Laboratory}
\acro{ROS}{Robot Operating System}
\acro{ROV}{Remotely Operated Vehicle}
\acro{SLAM}{simultaneous localisation and mapping}
\acro{GPS}{Global Positioning System}
\acro{EKF}{Extended Kalman Filter}
\acro{PF}{Particle Filter}
\acro{LBL}{Long Baseline}
\acro{USBL}{Ultra Short Baseline}
\acro{I-RRT*}{informed Rapidly-exploring Random Tree*}
\acro{RRT}{Rapidly-exploring Random Tree}
\acro{RRT*}{Asymptotic Optimal Rapidly-exploring Random Tree}
\acro{OMPL}{Open Motion Planning Library}
\end{acronym}

\def\BibTeX{{\rm B\kern-.05em{\sc i\kern-.025em b}\kern-.08em
    T\kern-.1667em\lower.7ex\hbox{E}\kern-.125emX}}
\begin{document}

\newcommand{\todo} [1] {\textcolor{red}{\textbf{TODO:} #1}}

\title{From market-ready ROVs to low-cost AUVs}

\author{Jonatan Scharff Willners\textsuperscript{b}, 
Ignacio Carlucho\textsuperscript{a},
Tomasz {\L}uczy{\'n}ski\textsuperscript{b}, 
Sean Katagiri\textsuperscript{b}, 
Chandler Lemoine\textsuperscript{a},
Joshua Roe\textsuperscript{b}, \\
Dylan Stephens\textsuperscript{a},
Shida Xu\textsuperscript{b}, 
Yaniel Carreno\textsuperscript{b}, 
\`Eric Pairet\textsuperscript{b}, 
Corina Barbalata\textsuperscript{a}, 
Yvan Petillot\textsuperscript{b},
Sen Wang\textsuperscript{b}\\



\textsuperscript{a }\textit{Department of Mechanical Engineering,} \textit{Louisiana State University}, Baton Rouge, USA \\
\{icarlucho, clemo17, dstep12, cbarbalata\}@lsu.edu \\

\textsuperscript{b }\textit{Institute of Sensors, Signals and Systems, Heriot-Watt University}, Edinburgh, UK \\
\{j.scharff\_willners, t.luczynski, s.katagiri, joshua.roe, sx2000, y.carreno, eric.pairet, y.r.petillot, s.wang\}@hw.ac.uk \\

}

\maketitle

\begin{abstract}

\acp{AUV} are becoming increasingly important for different types of industrial applications. The generally high cost of \acp{AUV} restricts the access to them and therefore advances in research and technological development. 
However, recent advances have led to lower cost commercially available \acp{ROV}, which present a platform that can be enhanced to enable a high degree of autonomy, similar to that of a high-end \ac{AUV}.
In this article, we present how a low-cost commercial-off-the-shelf \ac{ROV} can be used as a foundation for developing versatile and affordable \acp{AUV}. We introduce the required hardware modifications to obtain a system capable of autonomous operations as well as the necessary software modules. Additionally, we present a set of use cases exhibiting the versatility of the developed platform for intervention and mapping tasks.

\end{abstract}


\thispagestyle{plain} 
\pagestyle{plain} 
\section{Introduction}\label{sec:intro}
\let\thefootnote\relax\footnotetext{This work was supported by the EPSRC funded ORCA Hub (EP/R026173/1) and EU H2020 Programme under EUMarineRobots project (grant ID 731103).}

\acp{ROV} were first used for recovering torpedoes in the 1950s. 
Since then the technology has been adopted by a multitude of industries as their main approach for subsea operations. These robots have been previously utilized for tasks as diverse as, drilling for mineral and rock sampling \cite{ROVDrilling}, video survey of oil and gas platforms \cite{MCLEAN201966} and measuring the roughness of the seabed \cite{RoughnessROV}. 
However, continuous operation of \acp{ROV} can be costly as they require constant monitoring from an operator, who is connected with a tether to the \ac{ROV} from a support ship \cite{Robb2018}. Operating such a system can be extremely costly, with ships potentially being over \textsterling $50$M, and with reported costs of \textsterling $30$K per day \cite{TEAGUE2018333}. Additionally, this imposes a risk to operators who need to perform the launch or recovery of the \ac{ROV} on the deck, specially under rough weather.

Nevertheless, in the search for reducing cost and mitigating the risk to operators, research has been focusing on autonomous alternatives, such as \acp{AUV}. These vehicles, do not require constant monitoring from operators, and can perform different types of operations such as surveying \cite{TritonSurvey} or pipeline tracking \cite{PetillotDetection} on its own. 
Autonomous operations can therefore save a lot of time, as well as ease the deployment, which reduces the operative cost. 
It does however, depend on advanced and reliable software, which increases the development cost.

\begin{figure}
    \centering
    \includegraphics[width=\linewidth]{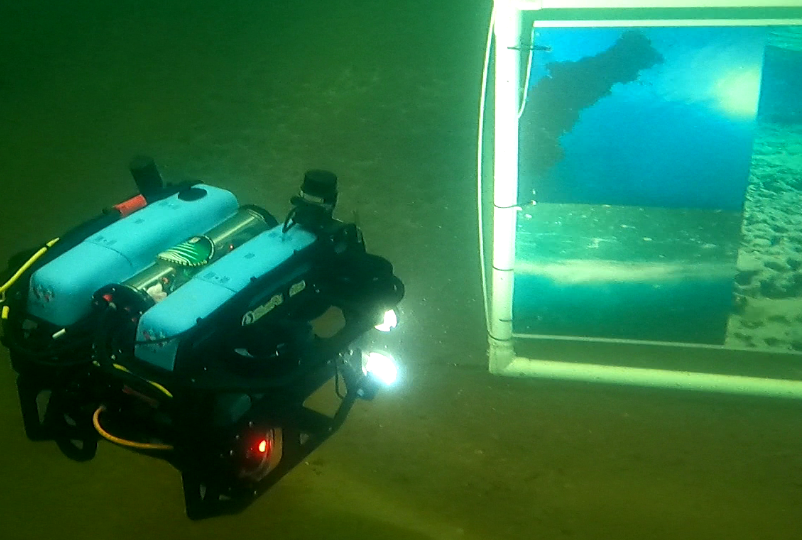}
    \caption{ The developed \ac{AUV} based on the BlueRov2 \ac{ROV}}
    \label{fig:dory}
\end{figure}

Previous research has shown different types of \acp{AUV} configuration and detailed how they can be constructed. 
Nessie-V, a $6$ \ac{DoF} research \ac{AUV}, was developed by \ac{OSL} at Heriot-Watt University to be used in research focused on inspection and autonomous missions  \cite{valeyrie2010nessie}.
Authors in \cite{Ribas2011} detail Girona~$500$, an \ac{AUV} with reconfigurable payload and propulsion system. In \cite{carreras2015testing}, Sparus~II \ac{AUV} is described which is a $5$ \ac{DoF} vehicle. Both Girona~$500$ and Sparus~II are powered by \ac{COLA2} \cite{Palomeras2012}, and commercially available with a base cost of around \$50,000. 
The modularity of these robots makes them suitable for multiple purposes including mapping and exploration.
In \cite{Ictiobo2019} an \ac{AUV} for acoustic imaging surveys, the Ictiobot-$40$,  was developed by the INTELYMEC group. However, the development of these systems from  ground up is extremely time-consuming and therefore not feasible for a large number of scientists.

Today, we see a shift, where cheaper \acp{ROV} are entering the market being affordable for hobbyists, small companies, and marine scientists. In this sense, the introduction of cheaper \ac{ROV} and sensors enables the possibility to combine \ac{COTS} products to create an \ac{AUV} at a lower cost and shorter time than building one from scratch. 
However, as expected with a lower price, there are trade-offs in terms of quality and capability. Most noticeable are the limited depth ratings, on-board sensors and the effect of the thrusters. The new affordable models are rarely rated for more than $100$m depth, they lack navigational sensors to reliably estimate their position, and due to their small size cannot handle well heavy sea-currents or waves. However, new \ac{COTS} \acp{ROV} can  be a good starting point in which an autonomous system can be built on top of, leading to savings both in time and cost. The cheaper \acp{ROV} can be improved with additional sensors for navigation, and improved software. Additionally, most of the applications of \ac{AUV}  are for usage in sheltered areas. Hence, the depth rating is sufficient and they are not exposed to the same amount of external disturbances as a work-class \ac{ROV} at open sea.

In this article we present how modern \ac{COTS} products can be combined to turn a \ac{ROV} into an \ac{AUV} for a fraction of the price when compared with a hovering capable \ac{AUV}. Our design is based on the BlueRov2 heavy configuration, shown in Fig. \ref{fig:dory}, a low-cost \ac{ROV} capable of working in depth of up to 100~m \cite{blue_robotics_2020}. 
We provide in this article a full description of the hardware modifications required for converting the \ac{ROV} to a fully autonomous \ac{AUV}, as well as the required software modules. 
Furthermore, we present real-time results for two different use cases i) Stereo vision with slam ii) Underwater payload manipulation. These results showcase the versatility of our proposed \ac{AUV} platform for performing a variable number of underwater missions, while costing a fraction of the price. 
 
The rest of the article is organized as follows. Section \ref{sec:background} presents the base platform used. Section \ref{sec:hardware} presents the hardware changes on the \ac{ROV}, while Section \ref{sec:software} presents the required software modifications. Section \ref{sec:experimental} introduces two use cases of the \ac{AUV}, and finally Section \ref{sec:Conclusions} presents some conclusions.

\section{Bluerov2 platform}
\label{sec:background}

The BlueRov2 is a \ac{COTS} \ac{ROV} produced by BlueRobotics \cite{blue_robotics_2020}. This small \ac{ROV} has a depth rating of $100$ meters, which provides excellent capabilities for exploring underwater environments. 
The small footprint of the vehicle and its weight under $12$~kg make it easy to transport and deploy.
Furthermore, the BlueRov2 comes equipped with a tether of up to 300m, which allows operators to control the vehicle remotely by means of a dedicated software package. The \ac{ROV} is internally powered, using a li-ion battery of $14.8$V and $18$Ah, providing ample energy autonomy. 

Different versions of the vehicle are commercially available, with a starting price of \textsterling $2.5$K. We based our design on the BlueRov2 heavy configuration, which comes with $8$ thrusters, arranged $4$ in vertical position and $4$ horizontally. This thruster configuration allows for a full $6$ \ac{DoF} control.   
Additionally, the BlueRov2 is equipped with a set of sensors that serve as feedback for basic control and exploration. The BlueRov2 has an \ac{IMU} (including accelerometer, gyroscope and compass), a depth sensor and a single-beam echosounder. Furthermore, it also has a front-facing tilting low-light camera, that can be enhanced with forward lights for underwater environments.
Moreover, the BlueRov2 also has current and voltage sensing capabilities, as well as leak detection circuits, extremely useful for secure underwater operations. 
With regards to the computing capabilities, the BlueRov2 has two computers on-board, the Pixhawk flight controller \cite{PIXHAWK} and a Raspberry Pi called a companion.

The BlueRov2 provides an exceptional starting platform for developing an autonomous solution. Based on the current system, in the following sections we will show what are the required hardware and software modifications necessary for transforming the \ac{COTS} \ac{ROV} into a fully functioning \ac{AUV}.

\section{Hardware}\label{sec:hardware}
Commercially available \acp{ROV} often have a very limited set of sensors, usually able to measure orientation and depth and cameras to aid the operator. To enable the platform to operate reliably in an autonomous fashion, additional hardware is often desired. In this section, we describe the necessary hardware modifications done on the BlueROV2 vehicle, to obtain autonomous capabilities.
We divide these modifications in four subsystems: i) Navigation sensors, ii) Perception sensors, iii) Data processing and decision-making system, iv) Communication system.

\subsection{Navigation Sensors}
There are a number of additional sensors that can support the quality of localisation, \eg \ac{USBL} can provide an absolute pose information by acoustic communication with an external transponder, \ac{IMU} can provide the orientation of the vehicle as well as angular velocity readings, and \ac{DVL} allows for fairly accurate linear velocity estimates. Inertial sensors and \ac{DVL} are sometimes combined in a single unit to provide \ac{DR}. The precision of the data provided by sensors used in \ac{DR} are usually related to the price. Designing a full localisation system at a low budget requires therefore a careful consideration as to which sensors should be selected.

In the interest of minimizing the cost, no \ac{USBL} was added to the systems used in our research. The \ac{ROV} used comes equipped with an \ac{IMU}, including a magnetometer, and a depth sensor. This gives absolute measurement for depth and orientation but no reliable information about movement in the x-y plane. To enable a full 6 \ac{DoF} pose estimation  we equipped the vehicle with a \ac{DVL}. We utilize the \ac{DVL} A50, from Waterlinked, due to its small footprint and accuracy on low seabeads. Additionally, this \ac{DVL} can be integrated with the ArduSub Companion. The \ac{DVL} was mounted to the lower section of the \ac{AUV} using a custom made 3D printed piece. The piece was engineered to effectively mount and stabilize the \ac{DVL} while not affecting the beams of the transducer. With this set of sensors on board the \ac{ROV}, sensor fusion can then be used for pose estimation. More details on this will be presented in Section \ref{sec:software}.

\subsection{Perception Sensors}

Underwater perception is usually performed either by acoustics (sonar) or optical (camera) sensors.
While cameras are widely available and the data is easy to understand for humans, the visibility in underwater environments is often limited due to water conditions \cite{DAI2020105947}, thus cameras are mainly effective at short distance (below a few metres). On the other hand, sonars are not affected by the visibility in the water and are therefore a more reliable choice when the visibility is limited. However, sonars are expensive and can be hard to interpret \cite{Villar2015}. Furthermore, the resolution of the data generated by sonars is also much lower, compared to cameras. Additionally, multiple cameras can be combined into stereo or multi-view setup, allowing for real-time 3D reconstruction, which can be a very valuable source of information.
We equipped the platform developed in this paper with a set of stereo cameras  with a custom-made sensor. Additionally, we include in the enclosures the required computational capabilities for processing the visual information.

When designing a stereo camera system, there are a number of factors to consider. Parameters like sensor type, lens, interface, and baseline selection can be modelled to match the requirements specific to the application \cite{Luczynski2019}. However, many factors may change in the development process, so two design principles are highlighted in the paragraphs below.

\begin{figure}
    \centering
    \includegraphics[width=\linewidth]{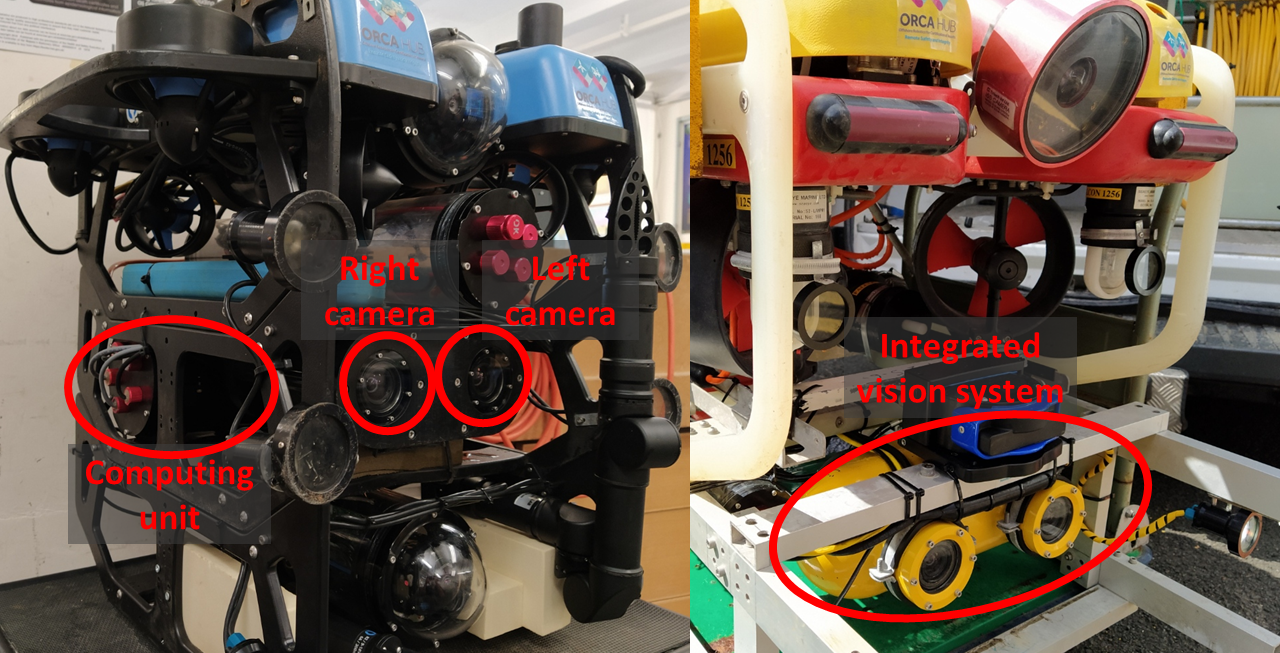}
    \caption{Two vision system designed according to different principles: left - a modular design with each component in smaller separate enclosure; right -  an integrated system with the cameras and all additional electronics are enclosed in a single housing.}
    \label{fig:cameras}
\end{figure}

First, we present the design of a single housing to incorporate all cameras. This allows for easy hardware integration, camera synchronization, and testing various hardware configurations. Processing units can also be easily added to process the images on-board. \figurename~\ref{fig:cameras} (right) shows our vision system designed according to this approach. This requires a dedicated underwater housing, which is difficult to design and produce, as well as much more expensive than typical cylinder-shaped enclosures. Furthermore, the relative pose of the cameras is fixed and cannot be changed. This approach is very useful during development but loses most of its advantages in the context of more mature systems, produced in larger numbers.

The second approach assumes that each camera is enclosed in a separate housing, and the processing units are also kept separately. This simplifies the construction of the enclosures and the components can be easily rearranged as necessary. \figurename~\ref{fig:cameras} (left) shows the system implemented in the \ac{AUV}, following these principles. The downside of this approach is that all components must be connected with underwater cables. This may become challenging as the cable and connectors need to rely on the protocol for data (and power) transfer, which might not always be commercially available.

\subsection{Data Processing and Decision-Making System}
To enable robust and reliable autonomy, the system needs to process large amounts of data, with as low latency as possible. This is a challenging task that grows with each software sub-module, especially in the context of algorithms responsible for data from the perception systems, which need to process a large amount of data at a high rate. Therefore, the selection of the embedded computers may heavily influence the performance of the system. Simpler tasks, such as running navigational sensors and handling the communication can be done on cheaper units ( \eg Raspberry Pi). Image data can be processed efficiently on FPGA, but these systems are not very flexible and increases the time for development, and are therefore better suited to finished and stable software. Therefore, since we are mostly working on the development level, we utilized embedded computers from the Nvidia Jetson family. Thanks to the included GPU, this boards are extremely effective for testing and implementing different vision algorithm in real-time. Additionally, with some care during the mounting stage the aluminium water tight enclosures from the BlueRobotics family can be utilised to provide efficient cooling to the processing units.

\subsection{Communication System} 
Ideally, an \ac{AUV} should work independently, with little or no communication with the operator. However, during the development process to ensure safety of operations, a constant communication link with the vehicle is required. This can be achieved in a few ways. Typically, the development process is conducted with a tether attached. This allows for continuous observation of the systems and streaming data to the surface, where more powerful processing units can be used. This allows the development of algorithms on the surface, before deciding on how much on-board processing power is needed on the embedded computers. The tether can also be moved to a communication buoy able to transfer data to the operator wirelessly. The vehicle can then move more freely and all the processing can be performed on-board, while at the same time, high-speed connections are available for data monitoring and debugging.

However, to fully become an autonomous vehicle the tether should preferably be removed completely as this can be a limitation for navigation, especially in structured and cluttered environments where the tether can get entangled. Without a tether, the only reliable source of communication is by using acoustic communication. There have been recent progress in optical communication which can provide a high data transfer rate but this has a limited range, can be highly affected by additional light sources (\eg the sun), and requires a direct line of sight. Acoustic communication is usually low bandwidth but capable of transmission for many kilometres.
In the developed platform, we removed the fixed tether connection present in the \ac{ROV}, and replaced it with an underwater connector from suburban marine \cite{suburbanmarine}. This connector utilizes the same penetrator size as the default Bluerov2, which allows for seamless integration, but can be easily connected or disconnected as required. 
The advantage of such a connector is that it allows a plug and play characteristic to the \ac{AUV}. 
In this way, the vehicle can perform tasks autonomously by disconnecting the tether completely, but if necessary, the tether can be easily connected and used, for example, in conjunction with a communication buoy to transmit mission information. This creates a versatile research platform that allows for rapid testing of algorithms and alternative communication techniques.

\section{Software}\label{sec:software}

In this section, we present 
the core software requirements necessary for obtaining an \ac{AUV} capable of following predefined waypoints. 
The developed system consists of the interpreter node, the pose estimation system, the control architecture and the waypoint pilot.
This general architecture will be referred as the navigation stack, upon which more intelligent and sophisticated autonomous behaviours and approaches can be built upon. Application examples of the navigation stack are presented in Section~\ref{sec:experimental}.

\begin{figure}
    \centering
    \includegraphics[width=0.45\textwidth]{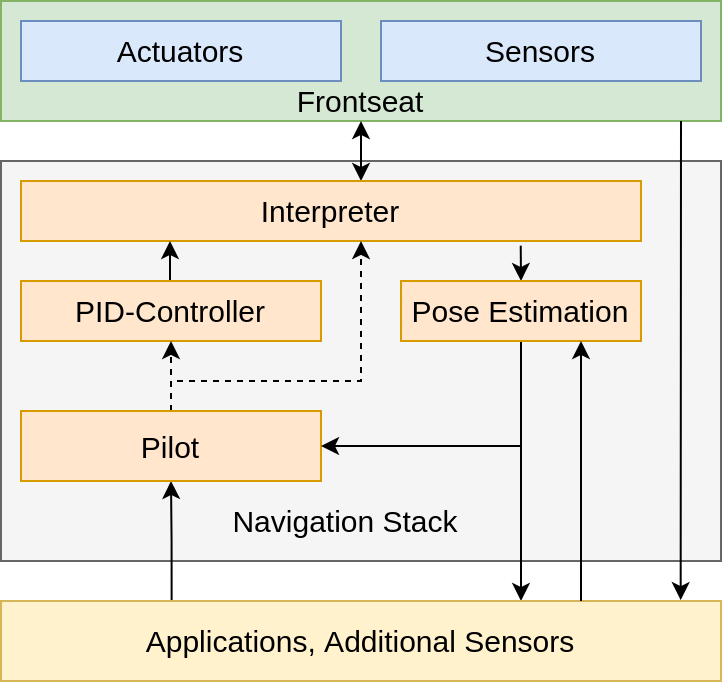}
    \caption{A navigation stack is used to connect the frontseat to the backseat to enable autonomous operation. If the frontseat can only be controlled using velocity/force commands a PID-controller is used for waypoint navigation.}
    \label{fig:navigation_stack}
\end{figure}

\subsection{Interpreter Node}
Commercially available \acp{ROV} usually come with a dedicated computer to supply the user with an interface for control and observation capabilities, \eg by sending velocity or thrust commands for controlling the movements of the \ac{ROV} or displaying sensor data as depth, heading and camera images. This computer is defined as a frontseat driver and usually has limited control capabilities. To enable more complex autonomous behaviours, a backseat driver can be used. The backseat driver is able to use the data from the \ac{ROV}'s sensors and command the \ac{ROV} through an interpreter node, which enables user-defined software to take control over the \ac{ROV}. The backseat driver can be deployed either as software on the frontseat's dedicated computer or as a separate computer connected to the frontseat. The architecture of the frontseat and the backseat can be seen \figurename~\ref{fig:navigation_stack}.

\ac{ROS} is being used in many robotics applications as the de facto standard for handling internal communication, offering an easy approach to a modular software system \cite{quigley2009}. Guided by this, we designed our system to leverage \ac{ROS} communications. \ac{ROS} allows for modular design and easy integration of additional modules, given that the communication layer is fast enough to transport the data as needed. All our vehicles are equipped with 1 GB network on board, which proved to be sufficient for performing the communicating and control tasks.

Since the Bluerov2 includes a frontseat computer, we include an interpreter node that can be used to handle the communication between the front and backseat. Furthermore, this allows the rest of the system to be platform agnostic and easily transferable between robots, allowing for faster deployment of algorithms in different platforms.  

\subsection{Pose Estimation System}
To enable autonomous navigation, the robot needs the ability to estimate its own position. For terrestrial robots, the access to GPS gives the possibility to measure an absolute position in the world at a reasonably high frequency. However, electromagnetic signals are absorbed in water. Hence, the same technology cannot be applied for an \ac{AUV} while submerged.
Instead \ac{DR} can be used to estimate the current position based on the internal sensors. To enable a full estimation in 6 \ac{DoF} the vehicle utilizes  the \ac{IMU} for orientation, a pressure sensor for depth and the \ac{DVL} for velocity estimation. In our navigation stack we use the Robot Localization Package from \ac{ROS} which fuses the sensor data through an \ac{EKF} \cite{Moore2016} to perform \ac{DR}. \ac{DR} is however based on the integration of data containing potential noise and biases, hence the error and uncertainty can therefore grow without bound. An alternative to \ac{DR} is to use natural features as references for estimating the pose using \eg visual \cite{Joshi2019} or acoustic \cite{Westman2018} \ac{SLAM}. A third approach is to use artificial landmarks such as acoustic beacons to communicate with the \ac{AUV} to perform \ac{LBL}, Moving-\ac{LBL} (MLBL) \cite{Fallon2010, Willners2019b}, or \ac{USBL} \cite{Ridao2011}. The different approaches can also be combined for improved pose estimation  \cite{Salavasidis2019, Vargas2021}.

In our implementation, we use additional odometry from a visual \ac{SLAM} node fused with the \ac{DR} generated by the frontseat, to improve the position estimate, further described in section \ref{section:slam}.
However, if the frontseat is able of performing position control based on its \ac{DR} and we have another pose estimation system, the two will drift apart over time. As we assume that the \ac{SLAM} odometry is more reliable over time, due to the fusion of more data, we continuously need to provide a transformation between the backseat estimation and the frontseat estimation. This enables the backseat to operate in one coordinate frame, while the frontseat operates in one based on the \ac{DR}.

\subsection{Control System}
To enable autonomous navigation between waypoints, a position and orientation controller is needed. We employ a generic navigation stack able to cope with two type of controllers in the frontseat: 1) position control \footnote{For ArduSub: https://www.ardusub.com/developers/dvl-integration.html} and 2) velocity/force control (through \eg a joystick). However, if the vehicle's frontseat is not endowed with this capability, 
the position can be controlled from the backseat using \eg a cascaded PID-controller \cite{Barbalata2015}.

\subsection{Waypoint Pilot}
A waypoint pilot is used to keep track of which waypoints, and in what order to visit them. The pilot serves as the interface that all high-level planning algorithms will use to control the robot. This node is able to improve the control of the robot indirectly by interpolating a path between waypoints as well as querying a path planner to find a collision-free path between two waypoints. As most hovering capable \acp{ROV} and \acp{AUV} are controllable freely in 3D, a geometric planner such as \ac{I-RRT*} \cite{Gammell2014}, implemented using \ac{OMPL} \cite{Sucan2012} can ensure collision-free trajectories within the known environment. If the vehicle needs to incorporate the kinematic constraints of the robot, planners as presented in \cite{Hernandez2018, pairet2020online, Willners2021} can be used.

\section{Experimental use cases}
\label{sec:experimental}

\begin{figure*}[!ht]
        \centering
           \subfloat[The \ac{AUV} performing SLAM. The green line shows ground truth (measured from an underwater motion tracking system) and the blue line shows the SLAM pose estimate.]{%
              \includegraphics[height=5cm,valign=c]{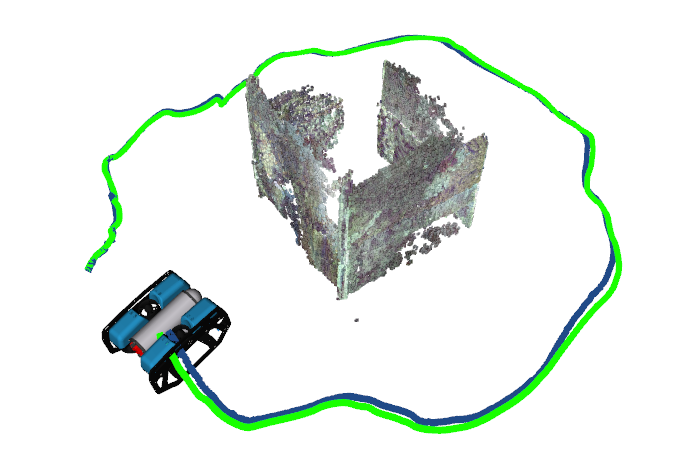}%
              \label{fig:slam}%
           } \hfil
           \subfloat[The AUV performing an intervention mission in an underwater environment using the Reach Alpha 5 manipulator.]{%
              \includegraphics[height=5cm,valign=c]{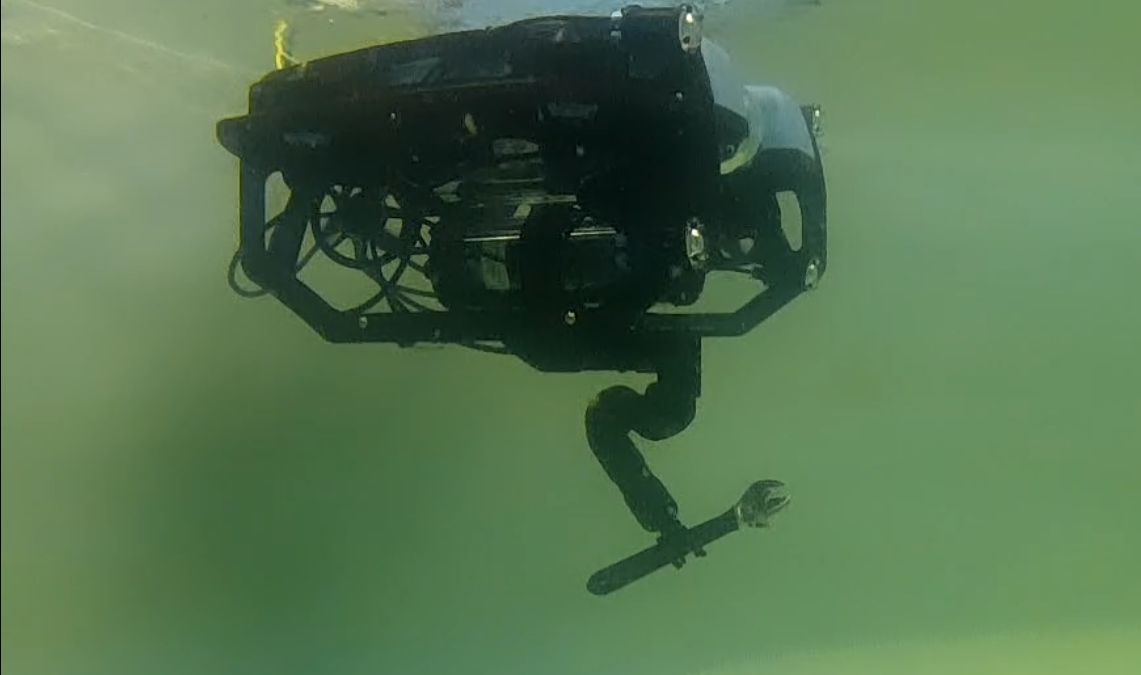}%
              \label{fig:arm}%
           }
           \caption{The developed \ac{AUV} has multiple use cases such as autonomous mapping and underwater intervention.}
           \label{fig:results}
\end{figure*}  

\subsection{Stereo Visual SLAM and Autonomous Inspection} \label{section:slam}

We include preliminary results that show how mapping can be performed with the developed system. The vehicle has been used with stereo vision ORB-SLAM \cite{Xu2021} with online extrinsic calibration of a \ac{DVL}, to incorporate the sensor measurements (velocity, depth and orientation) into the visual pose estimation. The approach was tested with ground truth as seen in \figurename~\ref{fig:slam}. 
The \ac{SLAM} system can be used by the navigation stack to improve the pose estimation by taking external features as reference in addition to \ac{DR}.

\subsubsection{Autonomous Robust Inspection}
An integrated \ac{SLAM} with active relocalisation for map-merging/loop-closure was deployed to test a robust underwater \ac{SLAM} system \cite{willners2021b}. The approach combines task-planning \cite{carrenoijcai2021} and viewpoint generation with the \ac{SLAM} system to endow the system with a map-merging procedure when visual tracking is lost. This is based on random sampling in a region around key-frames in the prior map, and simulating the sensor to find a location with a high probability to relocalise based on previously seen features.

\paragraph{High-Level Task Planning} AI planning solutions have shown promising results while solving complex missions in the underwater domain \cite{thompson2019review}, including environment's inspection. For the use case we present in this paper, the high-level task planner \cite{carrenoijcai2021} generates a \emph{plan}---sequence of actions that leads the robot from the initial to the final state where all goals are achieved---that allows the mapping of a structure amongst other actions. The plan actions are dispatched to the low-level system, including the hardware and software components previously discussed in this paper, for execution. Our system combines goal-based mission planning, based on a temporal planner \cite{benton2012temporal}, and a knowledge-based framework to achieve plans for dynamic problems. Therefore, this framework can adapt the initial plan to maintain robot operability when unexpected changes (not considered in the initial plan) occur. The knowledge-based framework encloses the  Situational Evaluation and Awareness (SEA) component. SEA is a failure solver which acts when failures occur, driving the robot from the failure state by proposing alternative behaviours or updating its knowledge (that translate on the generation of alternative plans). SEA evaluates the characteristics of failures to complete the global goals (consider in the initial plan) while introducing local goals that enable recovery from failure states. The possible mission failures embedded in SEA consider the tolerance analysis, anomaly and fault detection provided for past experiments using different approaches \cite{thompson2019review}. SEA creates a bridge between plan reasoning and execution to support system robustness when running for long periods. Combining the high-level planning approach and the low-level system, including a viewpoint planning module, allows implementing a robust structure mapping while mission survivability is guaranteed. When poor mapping is notified$^{*}$\footnote{$^{*}$The AUV losses track of the features it was using for localisation leading to the failure notification. In this situation, the robot will find new features and will start building a new map which is not optimal when mapping a structure.} by the low-level system, SEA advises the generation of a new plan that includes local goals for AUV relocalisation concerning the initial map. These local goals represent points provided by the viewpoint planning module, where it is highly expected that the AUV can merge the actual map with the original.

\subsection{Payload Manipulation}
Furthermore, we also include results for another use case that shows how the \ac{AUV} can be integrated with an underwater manipulator to perform intervention missions. The use case uses the Reach Alpha 5 \cite{reachalpha} underwater manipulator also integrated with \ac{ROS}, for manipulating unknown payloads in underwater environments.  

\begin{figure}
    \centering
    \includegraphics[width=\linewidth]{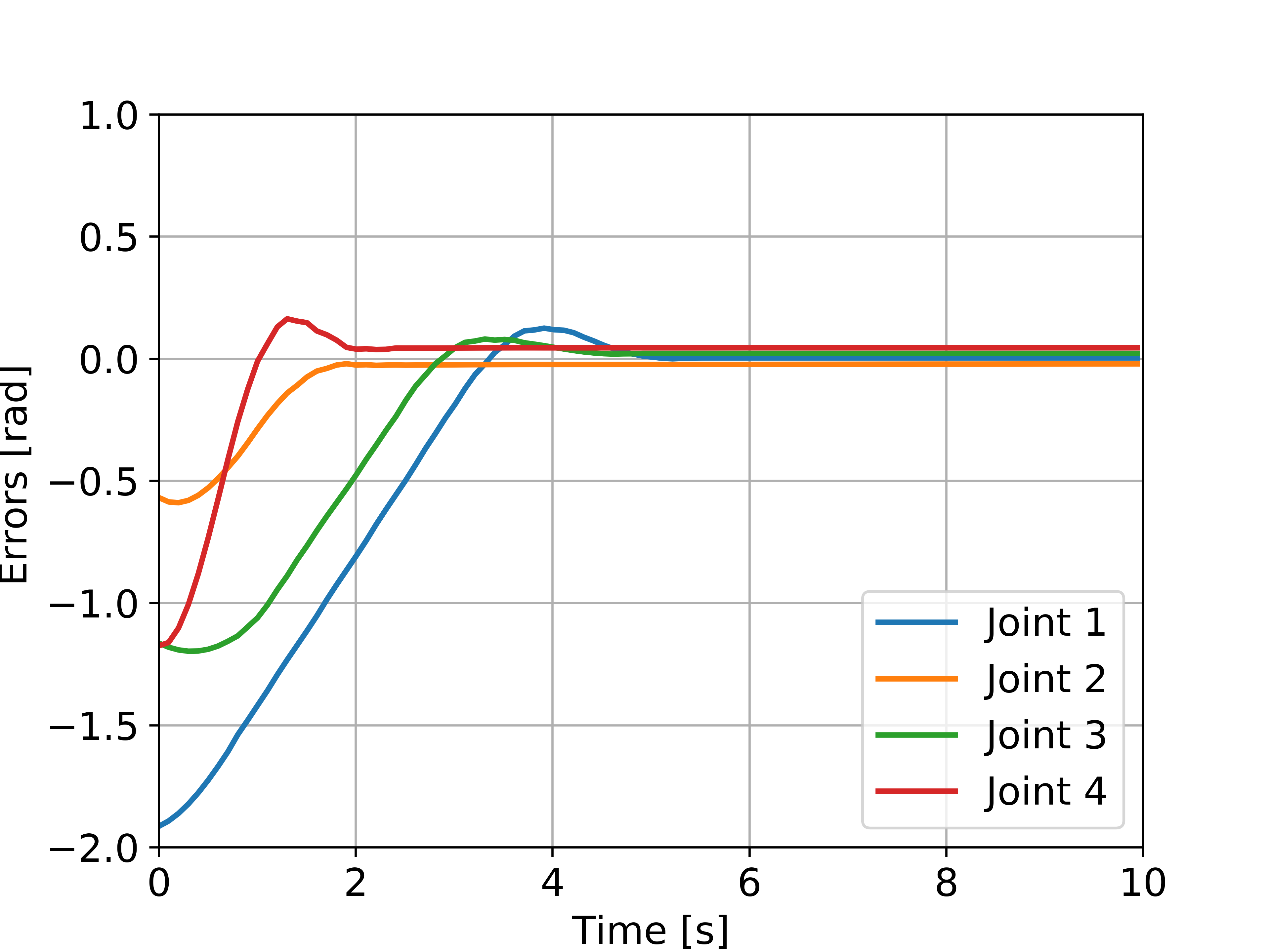}
    \caption{AdaNNMPC algorithm with arm holding a wrench with ${\mathbf{r}_t = [2.5, 2.0, 1.6, 2.0]}$ radians}
    \label{fig:adannmpcresult}
\end{figure}

In this test case, we focus on the low-level control of an underwater manipulator, mounted on an autonomous underwater vehicle, that was required to manipulate different payloads of unknown shape and size. Due to the highly nonlinear dynamics and the unknown hydrodynamics effects of the payloads, the performance of the control system degrade quite rapidly. To mitigate these effects, we developed a data-driven model predictive controller, based on neural networks. By utilizing a neural network, we were able to obtain a more accurate model, that inherently takes into consideration the environmental disturbances. Moreover, to account for the variations caused by the payloads, we enhance the control system with an online adaptive tuning strategy based on adaptive interaction theory. This adaptive mechanism takes into consideration the prediction window used in the optimal controller, allowing to take predictive tuning actions to improve the overall performance. We named this algorith the \ac{AdaNNMPC} \cite{CARLUCHO2021102726}. 
In Fig. \ref{fig:adannmpcresult} we show the results obtained when utilizing the \ac{AdaNNMPC} algorithm for a case study in which the arm is manipulating an unknown object, in this case a wrench, as shown in Fig. \ref{fig:arm}. The reference position ($\mathbf{r}_t $) in joint space is set as ${\mathbf{r}_t = [2.5, 2.0, 1.6, 2.0]}$ radians. It can be seen how the manipulator is able to reach the desired reference position in a short time, and with minimal overshoot, compensating for the variations caused by the payload. 

\section{Conclusion}\label{sec:Conclusions}
In this article, we have presented how a \ac{COTS} \ac{ROV} can be turned into an \ac{AUV} for fraction of the price, compared to commercially available \acp{AUV}. Likewise, the time saved by starting from an operating base platform compared to building it from scratch is large, 
and the economical benefit makes it the preferred approach for the majority of use cases. 
The paper presents the hardware changes as well as the software stack that constructs the core for an autonomous vehicle, capable of navigating without human intervention as well as providing a base on which to build more complex autonomous behaviours.
The vehicle used as an example is a BlueRov2, a low-cost 6 \ac{DoF} vehicle. 
However, the software, which is building on top of \ac{ROS} to enable autonomous capabilities, and parts of the vehicle has additionally been deployed on a Saab Seaeye Falcon$^{**}$\footnote{$^{**}$https://www.saabseaeye.com/solutions/underwater-vehicles/falcon} \acp{ROV}, demonstrating the generality of the proposed approach. 
We show a set of in-field experiments for mapping, planning, and interaction such as autonomous inspections, and manipulation, which has been carried out by the platform. The experiments shows the complex tasks the \ac{AUV} is capable at performing but for a fraction of the price of commercially available hovering capable \acp{AUV}.


\bibliographystyle{unsrt}
{\footnotesize \bibliography{references}}

\end{document}